\documentclass[twoside]{article}

\usepackage[sc]{mathpazo} 
\usepackage[T1]{fontenc} 
\usepackage{microtype} 

\usepackage[hmarginratio=1:1,top=32mm,columnsep=20pt]{geometry} 
\usepackage{multicol} 
\usepackage[hang, small,labelfont=bf,up,textfont=it,up]{caption} 
\usepackage{booktabs} 
\usepackage{float} 
\usepackage{hyperref} 

\usepackage{lettrine} 
\usepackage{paralist} 

\usepackage{abstract} 

\usepackage{titlesec} 
\renewcommand\thesection{\Roman{section}}
\titleformat{\section}[block]{\large\scshape\centering}{\thesection.}{1em}{} 

\usepackage{fancyhdr} 
\usepackage{amssymb}
\usepackage{epsfig}
\usepackage{amssymb}
\usepackage{url}
\usepackage{amsmath}
\usepackage{natbib}
\usepackage{amsthm}
\usepackage{booktabs}
\usepackage[lined,boxed,linesnumbered]{algorithm2e}
\usepackage{url}
\usepackage{alltt}
\usepackage{enumerate}
\usepackage{multirow}
\usepackage{setspace}
\usepackage{bm}
\usepackage{color}

\newcommand{\krnl}{\mathcal{K}}

\newcommand{\rx}{\mathtt{x}}
\newcommand{\sx}{\mathsf{x}}

\newcommand{\ry}{\mathrm{y}}

\newcommand{\vx}{\mathbf{x}}

\newcommand{\Id}{\mathcal{I}}

\newcommand{\Real}{I \! \! R}
\newcommand{\eye}{I \! \! I}

\newcommand{\sign}{\mathsf{sign}}

\newcommand{\nmathbf}{\bm}

\def\bfC{\nmathbf C}

\def\bfI{\nmathbf I}

\def\bfK{\nmathbf K}

\def\bfV{\nmathbf V}

\def\bfX{\nmathbf X}

\def\bfv{\nmathbf v}
\def\bfw{\nmathbf w}
\def\bfx{\nmathbf x}

\def\bfbeta   {\nmathbf \beta}

\def\bfmu     {\nmathbf \mu}

\def\bfLambda {\nmathbf \Lambda}

\def\bfSigma  {\nmathbf \Sigma}

\def\boldfacefake#1{\kern-4pt
    \hbox{ \mathsurround=0pt
    \hbox to 0.4pt{$#1$\hss}\hbox to 0.4pt{$#1$\hss}\hbox {$#1$}}}

\newcommand{\be}{\begin{eqnarray}}
\newcommand{\ee}{\end{eqnarray}}
\newcommand{\ba}{\begin{eqnarray*}}
\newcommand{\ea}{\end{eqnarray*}}

\newtheorem{theorem0}{Theorem}
\newtheorem{lemma0}{Lemma}
\newtheorem{remark0}{Remark}
\newtheorem{fact0}{Fact}
\newtheorem{example0}{Example}
\newtheorem{definition0}{Definition}
\newtheorem{corollary0}{Corollary}
\newtheorem{proposition0}{Proposition}
\newtheorem{algorithmY}{Algorithm}
\newtheorem{conjecture0}{Conjecture}

\newenvironment{theorem}{\begin{theorem0} \mbox{} }{\end{theorem0}}

\title{\vspace{-15mm}\fontsize{19pt}{10pt}
\selectfont\textbf{A Taxonomy of Big Data for Optimal Predictive Machine Learning and
Data Mining}} 

\author{
\textbf{Ernest Fokou\'e}\\
\normalsize Center for Quality and Applied Statistics \\ 
\normalsize Rochester Institute of Technology \\ 
\normalsize 98 Lomb Memorial Drive, Rochester, NY 14623, USA \\ 
\normalsize \href{mailto:ernest.fokoue@rit.edu}{ernest.fokoue@rit.edu} 
\vspace{-5mm}
}

\vspace{10mm}

\date{}

\begin{document}

\maketitle 

\thispagestyle{fancy} 


\begin{abstract}
\noindent Big data comes in various ways, types, shapes, forms and sizes. Indeed, almost all areas of science, technology, medicine, public health, economics, business, linguistics and social science  are bombarded by ever increasing flows of data begging to analyzed efficiently and effectively. In this paper, we propose a rough idea of a possible taxonomy of big data, along with some of the most commonly used tools for handling each particular category of bigness.
The dimensionality p of the input space and the sample size n are usually the main ingredients in the characterization of data bigness.  The specific statistical machine learning technique
used to handle a particular big data set  will depend on which category it falls in within the bigness taxonomy. Large p small n data sets for instance require a different set of tools from
the large n small p variety. Among other tools, we  discuss Preprocessing, Standardization, Imputation, Projection, Regularization,  Penalization, Compression, Reduction, Selection, Kernelization, Hybridization, Parallelization, Aggregation, Randomization, Replication, Sequentialization. Indeed, it is important to emphasize right away that the so-called no free lunch theorem applies here, in the sense that there is no universally
superior method that outperforms all other methods on all categories of bigness. It is also important to stress the fact that simplicity in the sense of Ockham's razor non plurality principle of parsimony tends to reign supreme when it comes to massive data. We conclude with a comparison of  the predictive performance of some of the most commonly used methods on a few data sets.
\end{abstract}


\section{Introduction}
We consider a dataset $\mathcal{D}=\{(\vx_1,y_1),(\vx_2,y_2),\cdots,(\vx_n,y_n)\},$ where $\vx_i^\top \equiv (\sx_{i1}, \sx_{i2},\cdots,\sx_{ip})$
denotes the $p$-dimensional vector of characteristics of the input space $\mathcal{X}$, and  $y_i$ represents the corresponding categorical response value
from the output space $\mathcal{Y}=\{1,\cdots,g\}$. Typically, one of the most basic ingredients in statistical data mining is the data matrix $\bfX$ given by
\begin{eqnarray}
\bfX=\left[\begin{array}{cccc}
x_{11} & x_{12} & \cdots & x_{1p}\\
x_{21} & x_{22} & \cdots & x_{2p} \\
\vdots & \ddots & \cdots & \vdots\\
x_{n1} & x_{n2} & \cdots & x_{np}\\
\end{array}\right].
\label{eq:data:mat:1}
\end{eqnarray}
Five aspects of the matrix $\bfX$ that are crucial to a taxonomy of massive data include: (i) The dimension $p$ of the input space $\mathcal{X}$, which simply represents the number of explanatory variables measured; (ii) The sample size $n$, which represents the number of observations (sites) at which the variables were measured/collected; (iii) The relationship between $n$ and $p$, namely the ratio $n/p$; (iv) The type of variables measured (categorical, ordinal, interval, count or real valued), and the indication of the scales/units of measurement; (v) The relationships among the columns of $\bfX$, namely whether or not the
columns are correlated (nearly linearly dependent). Indeed, as we will make clear later, massive data, also known as big data, come in various ways, types, shapes, forms and sizes. Different scenarios of massive data call upon tools and methods that can be drastically different at times. The rest of this paper is organized as follows:
 Section $2$ presents our suggested taxonomy for massive data based on a wide variety of scenarios.
 Section 3 presents a summary of the fundamental statistical learning theory along with some of the most
 commonly used statistical learning methods and their application in the context of massive data; Section 4 presents a comparison of predictive performances of some
popular statistical learning methods on a variety of massive data sets.
Section 5 presents our discussion and conclusion, along with some of the ideas we are planning to explore along the lines of the present paper.

\section{On the Diversity of Massive Data Sets}
\subsection*{\it Categorization of massiveness as a function of the input space dimensionality $p$}
Our idea of a basic ingredient for a taxonomy for massive data comes from
a simple reasoning. Consider the traditional multiple linear regression (MLR) setting with
$p$ predictor variables under the Gaussian noise. In a typical model space search
needed in variable selection, the best subsets approach fits $2^p - 1$ models
and submodels. If $p=20$, the space of linear models is of size
1 $\, million$. Yes indeed, one theoretically has to search a space of 1 $\, million$
models when $p=20$. Now, if we have $p=30$, the size of that space goes up
to 1 $\, billion$, and if $p=40$, the size of the model space goes up to 1 $\, trillion$,
and so on. {\it Our simple rule is that any problem with an input of more than $50$
variables is a massive data problem, because computationally searching
a thousand trillion is clearly a huge/massive task for modern day computers.}
Clearly, those who earn their keep analyzing inordinately large input spaces like the ones
inherent in microarray data ($p$ for such data is in the thousands)
will find this taxonomy somewhat naive, but it makes sense to us based
on the computational insights underlying it.
Besides, if the problem at hand requires the estimation of covariance
matrices and their inverses through many iterations, the $O(p^3)$ computational
complexity of matrix inversion would then require roughly $125000$ complexity
at every single iteration which could be quickly untenable computationally.
Now, it's clear that no one in their right mind decides to exhaustively
search a space of a thousand trillion models. However, this threshold gives
us somewhat of a point to operate from. From now on, any problem with
more than $50$ predictor variables will be a big data problem,
and any problem with $p$ exceeding $100$ will be referred to as a
massive data problem.

\subsection*{\it Categorization of massiveness as a function of the sample size $n$}
When it comes to ideas about determining how many observations one needs,
common sense will have it that the more the merrier. After all,
the more observations we have to close we are to the law of large numbers,
and indeed, as the sample size grows, so does the precision of our estimation.
However, some important machine learning methods like
Gaussian Process Classifiers, Gaussian Process Regression Estimators,
the Relevance Vector Machine (RVM),
Support Vector Machine (SVM) and just about all other kernel methods
operate in dual space and are therefore heavily dependent on
the sample size $n$.
The computational and statistical complexity of such methods is driven by the same size $n$.
Some of these methods like Gaussian Processes and
the Relevance Vector Machine require the inversion of
$n\times n$ matrices. As a result, such methods could easily be
computationally bogged down by too large a sample size $n$.
Now, how large is too large? Well, it takes $O(n^3)$ operations
to invert an $n \times n$ matrix. Anyone who works with
matrices quickly realizes that with modern-day computers,
messing around with more that a few hundreds in matrix inversion is not
very smart.  These methods can become excruciatingly (impractically)
slow or even unusable when $n$ gets ever larger. For the purposes
of our categorization, we set the cut-off at $1000$ and define as
{\it observation-massive} any dataset that have $n>1000$. Again, we derive this
categorization based on our observations on the
computational complexity of matrix inversion and its impact on
some of the state-of-the-art data mining techniques.

\subsection*{\it Categorization of massiveness as a function of the ratio $n/p$}
From the previous argumentation, we could say that
when $p>50$ or $n>1000$, we are computationally
in the presence of massive data. It turns out however
that the ratio $n/p$ is even more important to massive and learnability
than $n$ and $p$ taken separately. From experience,
it's our view that for each explanatory variable under study,
a minimum of $10$ observations is needed to have a decent analysis
from both accuracy and precision perspectives. Put in simple terms,
the number of rows must be at least $10$ times the number of
columns, specifically $n>10p$. Using this simple idea and the fact that
information is an increasing function of $n$, we suggest the following
taxonomy as a continuum of $n/p$.

\begin{table}[!h]
\centering
\begin{tabular}{lccc}
\toprule
    &   $\frac{n}{p} < 1$  & $1 \leq \frac{n}{p} < 10$  &  $\frac{n}{p} \geq 10$ \\
    & \textcolor{red}{\tt Information}  & \textcolor{blue}{\tt Information} &\textcolor{black}{\tt Information} \\
   &    \textcolor{red}{\tt Poverty} & \textcolor{blue}{\tt Scarcity} & \textcolor{black}{\tt Abundance}\\
   & \textcolor{red}{($n \lll p$)} & & \textcolor{black}{($n \ggg p$)} \\\toprule
\multirow{2}{*}{\tt $n > 1000$} &  {\tt Large $p$}, {\tt Large $n$} & {\tt Smaller $p$}, {\tt Large $n$} & {\tt Much smaller $p$}, {\tt Large $n$} \\
& \textcolor{red}{\tt A} & \textcolor{blue}{\tt B} & \textcolor{green}{\tt C}\\ \hline
\multirow{2}{*}{\tt $n \leq 1000$} &  {\tt Large $p$}, {\tt Smaller $n$} & {\tt Smaller $p$}, {\tt Smaller $n$} & {\tt Much smaller $p$}, {\tt Small $n$} \\
& \textcolor{magenta}{\tt D} & \textcolor{black}{\tt E} & \textcolor{yellow}{\tt F}\\
\bottomrule
\end{tabular}
\caption{In this taxonomy, \textcolor{red}{\tt A} and \textcolor{magenta}{\tt D} pose a lot of challenges.}
\end{table}

\section{Methods and Tools for Handling Massive Data}
\subsection*{\it Batch data vs incremental data production}
When it comes to the way in which the data is acquired or gathered, the traditionally assumed way is the so-called {\sf batch},
where all the data needed is available all at once. In state-of-the-art data mining however, there
are multiple scenarios where the data is produced/delivered in a sequential/incremental manner. This has prompted
the surge in the so-called {\sf online learning} methods. As a matter of fact, the {\sf perceptron}
learning rule, arguably the first algorithm that launched the whole field of
machine learning, is an {\sf online learning} algorithm. Online algorithms have the distinct advantage that
the data does not have to be stored in memory. All that is required in the storage of the
built model at time $t$. In the sense the stored model is assumed to have accumulated the structure of the underlying model.
Because of that distinct feature, one may think of using online algorithms even
when the whole data available. Indeed, when the sample size $n$ is so large that
the data cannot fit in the computer memory, one can consider building a learning method that
receives the data sequentially/incrementally rather than trying to load the whole data set into memory.
We shall refer to this aspect of massive data as sequentialization or incrementalization.
{\it Sequentialization} is therefore useful for both streaming data and massive data that
is too large to be loaded into memory all at once.

\subsection*{\it Missing Values and Imputation Schemes}
In most scenarios of massive data analytics, it is very common to be faced with
missing values. The literature on missing values is very large, and we will herein simply mention
very general guidelines. One of the first thing one needs to consider with missing values
is whether they are missing systematically or missing at random. The second important aspect is the rate of
missingness. Clearly, when we have abundance of data, the number of missing values is viewed differently.
Three approaches are often used to address missingness: (a) Deletion, which consists of deleting
all the rows that contain any missingness; (b) Central imputation, which consists of filling the
missing cells of the data matrix with central tendencies like mode, median or mean; (c)
model-based imputation using various adaptation of the ubiquitous Expectation-Maximization (EM) algorithm.

\subsection*{\it Inherent lack of structure and importance of preprocessing}
Sentiment analysis based on social media data from facebook and twitter, topic modelling
based on a wide variety of textual data, classification of tourist documents or even to be more general
the whole field of text mining and text categorization require the manipulation of inherently unstructured
data. All these machine learning problems are of great interest to end-users, statistical machine learning
practitioners and theorists, but cannot be solve without sometimes huge amounts of extensive pre-processing.
The analysis of a text corpus for instance never starts with a data matrix like the $\bfX$ defined in Equation
\eqref{eq:data:mat:1}. With these inherently unstructured data like text data,
the pre-processing often leads to  data matrices whose entries are frequencies of terms.
It's important to mention that term frequency matrices tend to contain many zeroes,
because a term deemed important for a handful of documents will tend not to appear in many other documents.
This content-sparsity can be a source of a variety of modelling problems.

\subsection*{\it Homogeneous vs Heterogeneous input space}
There are indeed many scenarios of massive data where the input space is {\sf homogeneous},
i.e. where all the variables are of the same type.
Audio processing, image processing and video processing all belong to a class of massive data where
all the variables are of the same type. There are however many other massive data scenarios where
the input space is made up of variables of various different types. Such heterogeneous input spaces
arise in fields like business, marketing, social sciences, psychology, etc  ... where one can have
categorical, ordinal, interval, count, and real valued variables gathered on the same entity. Such
scenarios call for {\sf hybridization}, which may take the form of combining two or more
data-type-specific methods in order to handle the heterogeneity of the input space.
In kernel methods for instance, if one has both {\tt textual} inputs and {\tt real valued} inputs,
then one could simply use a kernel $\mathcal{K} = \alpha \mathcal{K}_1 + (1-\alpha)\mathcal{K}_2$
that is the convex combination of two data-type-specific kernels, namely a string kernel $\mathcal{K}_1$
and a real valued kernel $\mathcal{K}_2$. {\it Hybridization} can also be used directly in modelling
through the use of combination of models.

\subsection*{\it Difference in measurement scale and the importance of transformation}
Even when the input space is homogeneous, it is almost always the case that the variables are measured
on different scales. This difference in scales can be the source of many modelling
difficulties. A simple way to address this {\it scale heterogeneity} is to perform
straightforward transformations that project all the variables onto the same scale.

{\it Standardization: } The most commonly used transformation is {\it standardization} which leads to all the
variables having zero mean and unit variance. Indeed, if $X_j$ is one of the variables
in $\mathcal{X}$ and we have $n$ observations $X_{1j}, X_{2j},\cdots,X_{nj}$, then the standardized
version of $X_{ij}$ is
$$
\tilde{X}_{ij} = \frac{X_{ij}-\bar{X}_j}{\sqrt{\sum_{i=1}^n{(X_{ij}-\bar{X}_j)^2}}},
\quad \text{where} \quad n\bar{X}_j = \sum_{i=1}^n{X_{ij}}
$$

{\it Unitization: } is another form of transformation commonly used. {\it Unitization} simply consists of transformation
the variables such that all take values in the unit interval $[0,1]$. The resulting input space is therefore
the {\it unit $p$-dimensional hypercube}, namely ${[0,1]^p}$.
With {\it unitization}, if $X_j$ is one of the variables
in $\mathcal{X}$ and we have $n$ observations $X_{1j}, X_{2j},\cdots,X_{nj}$, then the unitized
version of $X_{ij}$ is given by
$$
\tilde{X}_{ij} = \frac{X_{ij}-\min{X}_j}{\max{X}_j-\min{X}_j}.
$$

\subsection*{\it Dimensionality reduction and feature extraction}
Learning, especially statistical machine learning, is synonymous with dimensionality reduction.
Indeed, after data is gathered, especially massive data, nothing can be garnered in terms of insights
until some dimensionality reduction is performed to provide meaningful summaries revealing the
patterns underlying the data. Typically, when people speak of dimensionality reduction,
they have in mind the determination of some {\it intrinsic dimensionality} $q$ of the input space,
where $q \lll p$. There are many motivations for dimensionality reduction (a) achieve orthogonality
in the input space (b) eliminate redundant and noise variables, and as a result
perform the learning in a lower dimensional and orthogonal input space with the benefit
of variance reduction in the estimator. In practice, {\it lossy data compression techniques}
like principal component analysis (PCA) and
singular value decomposition (SVD) are the methods of choice for dimensionality reduction.
However, when $n \lll p$, most of these techniques cannot be directly used in their generic forms.

\subsection*{\it Kernelization and the Power of Mapping to Feature Spaces}
In some applications, like signal processing, it is always the case that $n \lll p$ in time domain. A ten seconds
audio track at a $44100 Hz$ sampling rate, generates a vector of dimension
$p=441000$ in time domain, and one typically has only few hundreds or maybe a thousand tracks for
the whole analysis. Typically image processing problems are similar in terms of dimensionality
with a simple face of size $640 \times 512$ generating a $p=327680$ dimensional input space.
In both these cases, it's impossible to use basic PCA or SVD, because $n \lll p$,
making it impossible to estimate the covariance structure needed in eigenvalue decomposition. One of the
solution to this problem is the use of the methods that operate in dual space, like
kernel methods. In recent years, {\it kernelization} has been widely applied and with tremendous
success to PCA, Canonical Correlation Analysis (CCA), Regression, Logistic Regression,
k-Means clustering just to name a few. Given a dataset with $n$ input vectors $\vx_i \in \mathcal{X}$
from some $p$ dimensional space, the main ingredient in kernelization is a bivariate
function $\krnl(\cdot,\cdot)$ defined on $\mathcal{X} \times \mathcal{X}$ and with values in $\Real$,
and the corresponding matrix of similarities $\bfK$ known as the Gram matrix and defined as
$$
\bfK=\left[\begin{array}{cccc}
\krnl(\bfx_1,\bfx_1) & \krnl(\bfx_1,\bfx_2) & \cdots & \krnl(\bfx_1,\bfx_n)\\
\krnl(\bfx_2,\bfx_1) & \krnl(\bfx_2,\bfx_2) & \cdots & \krnl(\bfx_2,\bfx_n)\\
\vdots & \ddots & \cdots & \vdots\\
\krnl(\bfx_n,\bfx_1) & \krnl(\bfx_n,\bfx_2) & \cdots & \krnl(\bfx_n,\bfx_n)\\
\end{array}\right].
$$

Crucial to most operations like kernel PCA is the  centered version of the Gram matrix given by
$$
\tilde{\bfK} = (\bfI_n - \eye_n)\bfK(\bfI_n - \eye_n) = \bfK -  \eye_n \bfK - \bfK \eye_n + \eye_n \bfK \eye_n,
$$
    where $\bfI_n \in \Real^{n \times n}$ and $\eye_n \in \Real^{n \times n}$ are both $n \times n$ matrices defined as
$$
{\eye}_n = \frac{1}{n}\left[\begin{array}{cccc}
1 & 1 & \cdots & 1\\
1 & 1 & \cdots & 1\\
\vdots & \ddots & \cdots & \vdots\\
1 & 1 & \cdots & 1\\
\end{array}\right] \quad \text{and} \quad
\bfI_n = \left[\begin{array}{cccc}
1 & 0 & \cdots & 0\\
0 & 1 & \cdots & 0\\
\vdots & \ddots & \cdots & \vdots\\
0 & 0 & \cdots & 1\\
\end{array}\right].
$$
The next step is to solve the eigenvalue problem
$$
\frac{1}{n}\tilde{K}\bfv_i = \lambda_i\bfv_i,
$$
where $\bfv_i \in \Real^n$ and $\lambda_i \in \Real$ for $i=1,\cdots,n$. In matrix form,
the eigenvalue problem is
$$
\frac{1}{n}\tilde{K} = \bfV \bfLambda \bfV^\top.
$$
In fact, basic PCA can be formulated in kernel form using the Euclidean inner product kernel
$\mathcal{K}(\vx_i, \vx_j) = \langle\vx_i,\vx_j\rangle = \bfx_i^\top\bfx_j$,
sometimes referred to as the vanilla kernel. If we center the data, i.e, such that $\sum_{i=1}^n{\vx_{ij}}=0$,
then the Gram matrix is
$$
\bfK = \left[\begin{array}{cccc}
\bfx_1^\top\bfx_1 & \bfx_1^\top\bfx_2 & \cdots & \bfx_1^\top\bfx_n\\
\bfx_2^\top\bfx_1 & \bfx_2^\top\bfx_2 & \cdots & \bfx_2^\top\bfx_n\\
\vdots & \ddots & \cdots & \vdots\\
\bfx_n^\top\bfx_1 & \bfx_n^\top\bfx_2 & \cdots & \bfx_n^\top\bfx_n\\
\end{array}\right]= \bfX\bfX^\top.
$$

Now, the covariance matrix is $\bfC=\frac{1}{n}\sum_{i=1}^n{\bfx_i\bfx_i^\top}= \frac{1}{n}\bfX^\top \bfX$, and PCA based on the covariance
is simply $\frac{1}{n}\bfX^\top \bfX \bfw_j = \lambda_j \bfw_j$ for $j=1,\cdots,p$ with $\bfw_j \in \Real^p$ and $\lambda_j \in \Real$.

\subsection*{\it Aggregation and the Appeal of Ensemble Learning}
It is often common in massive data that selecting a single model does not
lead the optimal prediction. For instance, in the presence of multicollinearity which is almost inevitable
when $p$ is very large, function estimators are typically unstable and of large variance.
The now popular {\it bootstrap aggregating} also referred to as {\it bagging} offers one way to
reduce the variance of the estimator by creating an aggregation of bootstrapped versions
of the base estimator. This is an example of ensemble learning, with the aggregation/combination
formed from equally weighted base learners.

{\it Bagging Regressors: } Let $\hat{\tt g}^{\tt (b)}(\cdot)$ be the $b$th bootstrap replication of the estimated base regression function $\hat{\tt g}(\cdot)$.
Then the {\it bagged} version of the estimator is given by
$$
\hat{{\tt g}}^{\tt (bagging)}(\vx) = \frac{1}{B}\sum_{b=1}^{B}{\hat{\tt g}^{\tt (b)}(\vx)}.
$$
If the base learner is a multiple linear regression model estimator $\hat{\tt g}(\vx)=\hat{\beta}_0 + \vx^{\top}{\hat{\bfbeta}}$, then the $b$th bootstrapped
replicate is  $\hat{\tt g}^{\tt (b)}(\vx)=\hat{\beta}_0^{(b)} + \vx^{\top}{\hat{\bfbeta}^{(b)}}$, and the bagged version
is
$$
\hat{{\tt g}}^{\tt (bagging)}(\vx) = \frac{1}{B}\sum_{b=1}^{B}{\left(\hat{\beta}_0^{(b)} + \vx^{\top}{\hat{\bfbeta}^{(b)}}\right)}
$$
{\it Bagging classifiers: } Consider a multi-class classification task with labels $\ry$ coming from $\mathcal{Y}=\{1,2,\cdots,m\}$ and predictor variables
 $\vx=(\rx_1,\rx_2,\cdots,\rx_q)^\top$  coming from a $q$-dimensional space $\mathcal{X}$.
 Let $\hat{\tt g}^{\tt (b)}(\cdot)$ be the $b$th bootstrap replication of the estimated base classifier $\hat{\tt g}(\cdot)$,
such that $(\hat{\ry})^{(b)} = \hat{\tt g}^{\tt (b)}(\vx)$ is the $b$th bootstrap  estimated class of $\vx$. The estimated response by bagging is obtained using the majority vote rule, which means the most frequent label throughout the $B$ bootstrap replications. Namely, $
\hat{{\tt g}}^{\tt (bagging)}(\vx) = \texttt{Most frequent label in } \hat{\tt C}^{\tt (B)}(\vx)$, where
$$
\hat{\tt C}^{\tt (B)}(\vx) = \bigg\{\hat{\tt g}^{\tt (1)}(\vx), \hat{\tt g}^{\tt (2)}(\vx), \cdots, \hat{\tt g}^{\tt (B)}(\vx)\bigg\}.
$$
Succinctly, we can write the bagged label of $\vx$ as
\begin{eqnarray*}
\hat{{\tt g}}^{\tt (bagging)}(\vx) = {\tt arg} \,\underset{\ry \in \mathcal{Y}}{\tt max}\left\{{\tt freq}_{\hat{\tt C}^{\tt (B)}(\vx)}(\ry)\right\}
= {\tt arg} \,\underset{\ry \in \mathcal{Y}}{\tt max}\left\{\sum_{b=1}^{B}{\left({\bf 1}_{\{\ry=\hat{\tt g}^{\tt (b)}(\vx)\}}\right)}\right\}.
\end{eqnarray*}

It must be emphasized that in general, ensembles do not assign equal weights to base learners in the aggregation. The general formulation
in the context of regression for instance is
\begin{eqnarray*}
\hat{{\tt g}}^{\tt (agg)}(\vx) =\sum_{b=1}^{B}{\alpha^{\tt (b)}\hat{\tt g}^{\tt (b)}(\vx)}.
\end{eqnarray*}
where the aggregation is often convex, i.e.  $\sum_{b=1}^B{\alpha^{\tt (b)}}=1$. \\

\subsection*{\it Parallelization}
When the computational complexity for building the base learner is high, using ensemble learning
techniques like bagging becomes very inefficient, sometimes to the point of being impractical. One way around this
difficulty is the use of parallel computation. In recent years, both R and Matlab have offered the capacity to
parallelize operations. Big data analytics will increasingly need parallelization as a way to speed up
computations or sometimes just make it possible to handle massive data that cannot fit into a single computer memory.

\subsection*{\it Regularization and the Power of Prior Information}
All statistical machine learning problems are inherently inverse problems, in the sense
that learning methods seek to optimally estimate an unknown generating function using empirical
observations assumed to be generated by it. As a result statistical machine learning problems
are inherently {\it ill-posed}, in the sense that they typically
violate at least one of Hadamard's three {\it well-posedness} conditions.
For clarity, according to Hadamard a problem is   {\it well-posed} if it fulfills the following three
conditions: {\it (a) A solution exists; (b) The solution is unique; (c) The solution is stable, i.e does not change
drastically under small perturbations.} For many machine learning problems, the first condition of well-posedness, namely
existence, is fulfilled. However, the solution is either not unique or not stable.
With large $p$ small $n$ for instance, not only is there a {\it multiplicity} of solutions but also
the instability thereof, due to the singularities resulting from the fact that $n \lll p$.
Typically, the regularization framework is used to isolate a feasible and optimal (in some sense)
solution. {\it Tikhonov}'s regularization is the one most commonly resorted to, and typically amounts to
a Lagrangian formulation of a constrained version of the initial problem, the constraints being the devices/objects
used to isolate a unique and stable solution.

\section{Statistical Machine Learning Methods for Massive Data}
We consider the traditional supervised learning task of pattern recognition with the goal of estimating a function $f$ that maps an input space $\mathcal{X}$
to a set of labels $\mathcal{Y}$. We consider the symmetric zero-loss $\ell(Y,f(X))= 1_{\{Y\neq f(X)\}}$, and the corresponding theoretical risk function
$$
R(f) = \mathbb{E}[\ell(Y,f(X))] = \int_{\cal{X} \times \cal{Y}}{\ell(y, f(\vx)) d P(\vx,y)}=\Pr[Y\neq f(X)].
$$
Ideally, one would like to find the universally best classifier $f^*$
that minimizes the rate $R(f)$ of misclassification, i.e.,
$$
 f^* = \underset{f}{\tt argmin}\Big\{\mathbb{E}[\ell(Y,f(X))]\Big\} = \underset{f}{\tt argmin}\Big\{\Pr[Y\neq f(X)]\Big\}.
$$
It is impossible in practice to find $f^*$, because that would require knowing the joint distribution of $(X,Y)$ which is usually unknown.
In a sense, $R(f)$, the theoretical risk, serves as a standard only and helps establish some important theoretical results in pattern recognition.
For instance, although in most practical problems one cannot effectively compute it,
it has been shown theoretically that the universally best classifier $f^*$ is the so-called Bayes classifier, the one obtained through the Bayes' formula
by computing the posterior probability of class membership as the discriminant function, namely,
$$
f^*(\vx) = {\tt class}^*(\vx)=\underset{j\in \{1,\cdots,g\}}{\tt argmax}{\left\{\Pr[Y=j|\vx]\right\}} =
\underset{j\in \{1,\cdots,g\}}{\tt argmax}{\left\{\frac{\pi_j p(\vx|y=j)}{p(\vx)}\right\}}.
$$
Assuming multivariate Gaussian class conditional densities with common covariance matrix $\Sigma$ and mean vectors $\bfmu_0$ and $\bfmu_1$,
the Bayes Risk, that is the risk associated to the Bayes classifier, is given by $R(f^*)=R^* = \Phi(-\sqrt{\Delta}/2)$ where $\Phi(\cdot)$ is the standard normal cdf and
$$
\Delta = (\bfmu_1-\bfmu_0)^\top \Sigma^{-1}(\bfmu_1-\bfmu_0).
$$
Once again, it is important to recall that this $R^*$ is not knowable in practice, and what typically happens is that, instead of seeking to
minimize the theoretical risk $R(f)$, experimenters focus on minimizing its empirical counterpart, known as the empirical risk.
Given an i.i.d sample $\mathcal{D}=\{(\vx_1,y_1),(\vx_2,y_2)\cdots,(\vx_n,y_n)\}$, the corresponding empirical risk is given by
$$
\hat{R}(f) = \frac{1}{n}\sum_{i=1}^n{1_{\{y_i\neq {f}(\vx_i)\}}},
$$
which is simply the observed (empirical) misclassification rate. It is re-assuring to know that fundamental results in statistical learning theory (See \cite{Vapnik:00:1}) establish that as the sample size goes to infinity, the empirical risk mimics the theoretical risk
$$
\underset{n\rightarrow \infty}{\lim}{\Pr[|\hat{R}(f)-{R}(f)|<\epsilon]} = 1.
$$
From a practical perspective,
this means that the empirical risk provides a tangible way to search the space of possible classifiers. Another crucial point
is the emphasis on the fact that even with this empirical risk, we still cannot feasibly search the universally
best function, for such a space would be formidably large. That's where the need to choose a particular function
class arises. In other words, instead of seeking an elusive universally best classifier, one simply proposes a plausible
classifier, possibly based on aspects of the data, then finds the empirical risk minimizer in that space,
and then, if the need arises maybe theoretically find out how the associated risk compares
to the Bayes risk.
One of the fundamental results in statistical learning theory has to do with the fact the minimizer of the
empirical risk could turn out to be overly optimistic, and lead to poor generalization performance. It is indeed
the case, that by making our estimated classifier very complex, it can adapt too well to the data at hand, meaning
very low in-sample error rate, but yield very high out of sample error rates, due to overfitting, the estimated
classifier having learned both the signal and the noise. In technical terms, this is referred to
as the bias-variance dilemma, in the sense that by increasing the complexity of the estimated classifier, the bias
is reduced (good fit all the way to the point of overfitting) but the variance of that estimator is increased.
On the other hand, considering much simpler estimators, leads to less variance but higher bias (due to underfitting,
model not rich enough to fit the data well). This phenomenon of bias variance dilemma, is particularly potent with
massive data when the number of predictor variables $p$ is much larger than the sample size $n$. One of the
main tools in the modern machine learning arsenal for dealing with this is the so-called regularization framework
whereby instead of using the empirical risk alone, a constrained version of it, also known as the regularized or penalized
version is used.
$$
\hat{R}_{\tt reg}(f) =  \hat{R}(f) + \lambda \|f\|_{\cal{H}}= \frac{1}{n}\sum_{i=1}^n{1_{\{y_i\neq {f}(\vx_i)\}}} + \lambda \|f\|_{\cal{H}},
$$
where $\lambda$ is referred to as the tuning (regularization) parameter, and $\|f\|_{\cal{H}}$ is some measure of the complexity of $f$ with
the class $\cal{H}$ from which it is chosen. It makes sense that choosing a function $f$ with a smaller value of $\|f\|_{\cal{H}}$
helps avoid overfitting. The value of $\lambda \in [0,+\infty)$, controls the trade-off between bias (goodness of fit), and
function complexity (which is responsibility for variance). Practically though, it may still be hard to even explore the theoretical properties of a given
classifier and compare it to the Bayes risk, precisely because, methods typically do not directly act on the zero-one loss function, but instead
use at best surrogates of it. Indeed,  within a selected class $\cal{H}$ of potential classifiers, one typically
chooses some loss function $\ell(\cdot,\cdot)$ with some desirable properties like
smoothness and/or convexity (this is because one needs at least to be able to build the desired classifier),
and then finds the minimizer of its regularized version, i.e.,
$$
\hat{R}_{\tt reg}(f) = \frac{1}{n}\sum_{i=1}^n{\ell(y_i, {f}(\vx_i))} + \lambda \|f\|_{\cal{H}}.
$$
Note that $\lambda$ stills controls the bias-variance trade-off as before. Now, since the loss function typically chosen
is not the zero-one loss on which the Bayes classifier (universally best) is based, there is no guarantee that the best in the selected
class $\cal{H}$ under the chosen loss function $\ell(\cdot,\cdot)$ will mimic $f^*$. As a matter of fact, each optimal classifier
from a given class $\cal{H}$ will typically perform well if the data at hand and the generator from which ut came, somewhat accord
with the properties of the space $\cal{H}$. This remark is probably what prompted the famous so-called no free lunch theorem, herein stated informally.

\begin{theorem}{(\tt No Free Lunch)}
There is no learning method that is universally superior to all other methods on all datasets.
In other words, if a learning method is presented with a data set whose inherent patterns
violate its assumptions, then that learning method will under-perform.
\end{theorem}

The above no free lunch theorem basically says that there is no such thing as a universally superior learning method
that outperforms all other methods on all possible data, no matter how sophisticated the method may appear to be. Indeed,
it is very humbling to see that some of the methods deemed somewhat simple sometimes hugely outperform the most sophisticated
ones when compared on the basis of average out of sample (test) error. It is common practice in data mining and machine
learning in general, to compare methods based on benchmark data, and empirical counterparts of the theoretical predictive measures,
often computed using some form of re-sampling tool like the bootstrap or cross-validation.
Massive data, also known as big data, come in various types, shapes, forms and sizes. The specific statistical machine learning technique
used to handle a particular massive data set depends on which aspect of the taxonomy  it falls in. Indeed,
it is important to emphasize that the no free lunch theorem applies more potently here, in the sense that there is no panacea that universally
applies to all massive data sets. It is important however to quickly stress the fact that simplicity in the sense of Ockham's razor
non plurality principle of parsimony tends to reign supreme when it comes to massive data. In this paper, we propose a rough idea of a possible taxomony
of massive data, along with some of the most commonly used tools for handling each particular class of massiveness. In this paper,
we consider a few datasets of different types of massiveness, and we demonstrate through computational results that
the no free lunch applies as a stronger as ever. We typically consider some of the most commonly used pattern recognition techniques,
from those that are most simple and intuitive to some that are considered sophisticated and state-of-the-art, and we show that
the performances vary sometimes drastically from data to data. It turns out, as we will show, that depending
on the type of massiveness, some methods cannot even be used. We also provide our taxonomy of massiveness along with different approaches to dealing
with each case. See \cite{Vapnik:00:1} and \cite{GuoY:2005:1}

\subsection*{\it Linear Discriminant Analysis}
Under the assumption of multivariate normality of the class conditional densities with equal covariance matrices, namely $(\vx|y=j) \sim {\tt MVN}(\bfmu_j, \Sigma)$,
or specifically,
$$
p(\vx|y=j) = \frac{1}{(2\pi)^{p/2}|\Sigma|^{1/2}}\exp\left\{-\frac{1}{2}(\vx-\bfmu_j)^\top\Sigma^{-1}(\vx-\bfmu_j)\right\},
$$
a classifier with excellent desirable properties is Linear Discriminant Analysis (LDA)
classifier, which, given any new point $\vx$,  will estimate the corresponding class $Y$ as
\begin{eqnarray*}
\hat{Y}_{\tt \tiny LDA}=\hat{f}_{\tt \tiny LDA}(\vx) = 1_{\left\{\hat{\beta}_0+\hat{\bfbeta}^\top\vx>0\right\}} = 1_{\{\delta_1(\vx)>\delta_0(\vx)\}}
\end{eqnarray*}
where
$$
\delta_j(\vx) =  \vx^\top  \bfSigma^{-1}\bfmu_j  -\frac{1}{2}\bfmu_j^\top\bfSigma^{-1}\bfmu_j  + \log \pi_j
$$
or
$$
\hat{\bfbeta} = \hat{\Sigma}^{-1}(\hat{\bfmu}_1-\hat{\bfmu}_0)
\quad
\text{and} \quad
\hat{\beta}_0 = -\frac{1}{2}(\hat{\bfmu}_1+\hat{\bfmu}_0)^\top\hat{\Sigma}^{-1}(\hat{\bfmu}_1-\hat{\bfmu}_0) + \log\frac{\hat{\pi}_1}{\hat{\pi}_0}
$$
where using the indicators $z_{ij}=I(y_i=j)=1_{\{y_i=j\}}$, the estimated prior probabilities of class membership is given by $\hat{\pi}_j = ({n_j}/{n})= ({1}/{n})\sum_{i=1}^n{z_{ij}}$, the sample mean vectors $\bfmu_j$ are given by $\hat{\bfmu}_j = ({1}/{n_j})\sum_{i=1}^n{z_{ij}\vx_i}$ and the
sample covariance matrices $S_j = ({1}/{(n_j-1)})\sum_{i=1}^n{z_{ij}(\vx_i-\hat{\bfmu}_j)(\vx_i-\hat{\bfmu}_j)^\top}$, so that
the sample pooled covariance
\begin{eqnarray*}
{\hat\Sigma} = (1/n)\sum_{j=0}^1{\sum_{i=1}^n{z_{ij}(\vx_i-\hat{\bfmu}_j)(\vx_i-\hat{\bfmu}_j)^\top}}
\end{eqnarray*}
The estimation of $\Sigma$ is clearly central to the use of Linear Discriminant Analysis. However, when $n \lll p$,
$\hat{\Sigma}$ is ill-defined at best. One of the approaches to dealing with this drawback is the use of regularization which essentially consists
of adding a jitter to the diagonal of $\hat{\Sigma}$. For $\alpha \in (0,1)$ or alternatively for $\lambda \in (0,\infty)$, the regularized version
$\hat{\Sigma}$, namely $\hat{\Sigma}_{\tt reg}$ can be obtained
$\hat{\Sigma}_{\tt reg} = (1-\alpha)\hat{\Sigma} + \alpha I$ or $\Sigma_{\tt reg} = \hat{\Sigma} +  \lambda I$ or even $\Sigma_{\tt reg} = \lambda\hat{\Sigma} +  I$. \cite{Friedman:1989:1} provides one of the earliest full treatment of this
approach to singularity in LDA. \cite{GuoY:2005:1} give a very compelling application of this kind of regularization to microarray data.

\subsection*{\it Nearest Neighbors Methods}
The so-called $k$-Nearest Neighbors approach to learning computes the estimated response  $\hat{Y}_{\tt kNN}^*$ of a new  point $\vx^{*}$  as follows:
first set the neighborhood size $k$; then choose a distance measure $d(\cdot, \cdot)$ defined on $\mathcal{X}\times\mathcal{X}$; then compute $d_i^* = d(\vx^*,\vx_i),\,\, i=1,\cdots,n$, then rank all the distances $d_{i}^*$ in increasing order
$d_{(1)}^* \leq d_{(2)}^*\leq \cdots \leq d_{(k)}^* \leq d_{(k+1)}^*\leq \cdots \leq d_{(n)}^*$. Then specify
$\mathcal{V}_k(\vx^*) = \big\{\vx_i: \,\, d(\vx^*,\vx_i) \leq d_{(k)}^*\big\}$, the size $k$ neighborhood of $\vx^{*}$, made up of the $k$ points in $\mathcal{D}$ that are closest to $\vx^{*}$ according to $d(\cdot,\cdot)$.
The estimate response is the $\textit{Most frequent label in } \mathcal{V}_k(\vx^*)$, specifically
\begin{eqnarray}
\hat{Y}_{\tt kNN}^* = \hat{f}_{\tt kNN}(\vx^*) =  \underset{j \in \{1,\cdots,g\}}{{\tt arg}{\tt max}} \left\{p_j^{(k)}(\vx^*)\right\}
\label{eq:kNN:class:1}
\end{eqnarray}
where
\begin{eqnarray}
p_j^{(k)}(\vx^*) = \frac{1}{k}\sum_{i=1}^n{I(Y_i = j)I(\vx_i \in \mathcal{V}_k(\vx^*))}
\label{eq:kNN:class:2}
\end{eqnarray}
estimates the probability that $\vx^*$ belongs to class $j$ based on $\mathcal{V}_k(\vx^*)$.  Indeed, $\displaystyle p_j^{(k)}(\vx^*)$ can be thought of as a rough estimate of $\pi_j(\vx^*)=\Pr[Y^*=j|\vx^*]$, the posterior probability of class membership of $\vx^*$,ie
$p_j^{(k)}(\vx^*) = \approx \widehat{\pi_j(\vx^*)}$. See \cite{HTF:1} and \cite{CFZ:1}.
kNearest Neighbors (kNN) essentially performs classification by voting for the most popular response among
the $k$ nearest neighbors of $\vx^*$. In a sense, {\tt kNN} provides the most basic form of nonparametric classification
Thanks to the fact that the estimated response $\hat{Y}_{\tt kNN}^*$ for $\vx^*$ is  - at least - a crude nonparametric estimator of Bayes classifier's response,
which somewhat justifies (or at least explains) the great interest in kNN.
Typical distances used in practice include the Euclidean distance $d(\vx_i, \vx_j) = \|\vx_i-\vx_j\|_2 =  \left(\sum_{\ell=1}^p{(\sx_{i\ell}-\sx_{j\ell})^2}\right)^{1/2}$,
and the Manhattan distance $d(\vx_i, \vx_j) = \|\vx_i-\vx_j\|_1 =  \sum_{\ell=1}^p{|\sx_{i\ell}-\sx_{j\ell}|}$.
One of the appeals of the $k$-Nearest Neighbors approach in both classification and regression lies in the fact it seamlessly applied to
data of any type, as long as a valid and appropriate distance can be defined. For instance, with binary data where each $\vx_i^\top = (\sx_{i1}, \cdots,\sx_{ip}) \in \{0,1\}^p$ is a $p$-dimensional vector of binary (indicator) values, one could use the Hamming distance or the very useful
Jaccard distance defined as $d_J(\vx_i,\vx_j) = 1- J(\vx_i,\vx_j)$, where, $J(\vx_i,\vx_j)$ is the Jaccard similarity index, defined
by
$$
J(\vx_i,\vx_j) = \frac{\vx_i^\top \vx_j}{\vx_i^\top\vx_i+\vx_j^\top\vx_j-\vx_i^\top \vx_j}.
$$
Clearly, $|J(\vx_i,\vx_j)| \leq 1$, with maximum value of $1$ attained at $J(\vx_i,\vx_i)$ and minimum value of $0$ corresponding to two vectors with no matching $1$ values. Since the fundamental building block of {\tt kNN} is the distance measure, one can easily perform classification beyond the traditional setting where the predictors are numeric. For instance, classification with kNN can be readily performed on indicator attributes $\vx_i = (\rx_{i1}, \cdots, \rx_{ip})^\top \in \{0,1\}^{p}$.
{\tt kNN} classifiers are inherently naturally multi-class, and are used extensively in applications such as image processing, character recognition and general pattern recognition tasks.

When it so happens as it often does that the input space $\cal{X}$ is nonhomogeneous, a typical approach to nearest neighbors pattern recognition
consists of defining a distance that is the direct or convex sum of the type-specific variables. For instance, if there is a group of numeric variables
and a group of categorical variables, one could use the distance $d_1(\cdot,\cdot)$ for the first group, and $d_1(\cdot,\cdot)$ for the second group, and then
either use the direct sum $d(\vx_i,\vx_j) = d_1(\vx_i,\vx_j) + d_2(\vx_i,\vx_j)$ or the convex sum $d(\vx_i,\vx_j) = \alpha d_1(\vx_i,\vx_j) + (1-\alpha)d_2(\vx_i,\vx_j)$ for some $\alpha \in (0,1)$.

Nearest Neighbors Algorithms are extremely popular in Statistical Data Mining and Machine Learning probably due to the fact that
they offer the simplest and most flexible form of nonparametric predictive analytics.
In this paper, we explore various aspects of predictive analytics in regression and pattern recognition (classification) using kNearest Neighbors.
The complexity of the underlying pattern in k-Nearest Neighbors Analysis is controlled by the size $k$ of the neighborhood. We show how Cross validation and other forms of re-sampling can be used to determine the optimal value $k$ that helps achieve bias-variance trade-off and thereby good generalization (low prediction error).

\subsection*{\it Support Vector Machines}
Consider a response variable taking values in $\{-1,+1\}$ and a regularized empirical risk functional given by
$$
\hat{R}_{\tt reg}(\beta_0,\bfbeta) = \frac{1}{n}\sum_{i=1}^n{(1-y_i(\beta_0+\bfbeta^\top\vx_i))_{+}} + \frac{\lambda}{2}\sum_{j=0}^p{|\beta_j|^2},
$$
where $\lambda \in \Real_+^*$ is the {\it regularization} (tuning) hyperparameter, and the function $(u)_+=\max(0,u)$ is used to define the {\it hinge loss} $$
\ell(y_i, f(\vx_i)) = (1-y_i(\beta_0+\bfbeta^\top\vx_i))_{+} =  \left\{\begin{array}{ll} 0 & \, \text{if} \,\, 1-y_i(\beta_0+\bfbeta^\top\vx_i) < 0 \\
1-y_i(\beta_0+\bfbeta^\top\vx_i) & \, \text{if} \,\, 1-y_i(\beta_0+\bfbeta^\top\vx_i) \geq 0
\end{array}\right.
$$
Solving the optimization problem
$$
(\hat{\beta}_0,\hat{\bfbeta}^\top)^\top = \underset{(\beta_0,\bfbeta^\top)^\top\in \Real^{p+1}}{\tt argmin}\left\{\frac{1}{n}\sum_{i=1}^n{(1-y_i(\beta_0+\bfbeta^\top\vx_i))_{+}} + \frac{\lambda}{2}\sum_{j=0}^p{|\beta_j|^2}\right\}
$$
yields the linear support vector machine (SVM) binary classifier whose predicted response is
$$
\hat{f}_{\tt svm}(\vx) = \sign\left(\hat{\beta}_0+\hat{\bfbeta}^{\top}\vx\right)
$$
The above solution assumes that the decision boundary between the two classes is a hyperplane. However,
it often happens that the decision boundary is not linear. {\it Kernelization} is the standard approach to handling the
nonlinearity of the decision boundary in support vector machine classification.  The kernel version of the SVM classifier
is given by
$$
\hat{f}_{\tt svm}(\vx) = \sign\left(\sum_{i=1}^n{y_i\hat{\alpha}_i\mathcal{K}(\vx,\vx_i)+ \hat{\beta}_0}\right),
$$
where $\mathcal{K}(\cdot,\cdot)$ is a bivariate function called {\it kernel} defined on $\mathcal{X} \times \mathcal{X}$, and used
to measure the similarity between two points in observation space. The $\hat{\alpha}_i$'s
are determined via quadratic programming optimization, with the nonzero $\hat{\alpha}_i$'s corresponding to the so-called
support vectors. One of the most general and most commonly used kernel in the context of pattern recognition is the Gaussian Radial Basis Function kernel given by
$$
\mathcal{K}(\vx_i, \vx_j) = \exp\Bigg(-\frac{1}{2}\frac{\|\vx_i-\vx_j\|_2^2}{\tau^2}\Bigg).
$$
The second most commonly used kernel is the Laplace radial basis function kernel defined as
$$
\mathcal{K}(\vx_i, \vx_j) = \exp\Bigg(-\frac{1}{2}\frac{\|\vx_i-\vx_j\|_1}{\tau}\Bigg).
$$
There are many other kernels like the polynomial kernel $\mathcal{K}(\vx_i, \vx_j) = ({\tt scale}\langle\vx_i,\vx_j\rangle+{\tt offset})^{\tt degree}$,
and others, see \cite{CFZ:1}, \cite{Bousquet:03:1}, \cite{Tipping:01:1}, \cite{Vapnik:00:1} among others. When it so happens that the input space
is nonhomogeneous, one could remedy by defined a {\it hybrid} that is a linear convex combination of other dat-type specific kernels, namely
$$
\mathcal{K}(\vx_i,\vx_j) = \alpha \mathcal{K}_1(\vx_i,\vx_j) + (1-\alpha) \mathcal{K}_2(\vx_i,\vx_j)
$$
It can therefore be said that the Support Vector Machine has the potential to use massive data tools like Regularization, Kernelization,
Hybridization. In fact, we will see later that even {\it aggregation}/{\it combination} will be using in both the {\it bagging} and {\it boosting}
SVM classifiers.

\subsection*{\it Logistic Regression}
Arguably one of the most widely used pattern recognition machines, logistic regression is indeed used in the context of
massive data with both regularization and kernelization frequently resorted to in a variety of scenarios.
Using the traditional $\{0,1\}$ indicator labelling, the empirical risk for the binary linear logistic regression model is given by
$$
\hat{R}(\beta_0,\bfbeta)=-\frac{1}{n}\sum_{i=1}^n\left\{y_i(\beta_0+\bfbeta^\top\vx_i)-\log\left[1+\exp\left\{(\beta_0+\bfbeta^\top\vx_i)\right\}\right]\right\}.
$$
However, if we use the labelling $\{-1,+1\}$ as with SVM, the empirical risk for the linear logistic regression model is given by
$$
\hat{R}(\beta_0,\bfbeta)=\frac{1}{n}\sum_{i=1}^n{\log\left[1+\exp\left\{-y_i(\beta_0+\bfbeta^\top\vx_i)\right\}\right]}.
$$
One of the most common extensions of this basic formulation of logistic regression is the use of the regularized version of the empirical risk with $\ell_1$ norm
on the space of the coefficients $\beta_j$'s. This so-called LASSO logistic regression achieves both shrinkage and variable selection in situations where
the data contains redundant and/or noise variables.
$$
\hat{R}_{\tt reg}(\beta_0,\bfbeta) = \frac{1}{n}\sum_{i=1}^n{\log\left[1+\exp\left\{-y_i(\beta_0+\bfbeta^\top\vx_i)\right\}\right]} + \lambda\sum_{j=0}^p{|\beta_j|}
$$
An even more powerful extension of the logistic regression model comes with the use of kernels to capture the nonlinearity of underlying decision boundary of the classifier. Using kernels as defined earlier, the regularized empirical risk for the kernel logistic regression is given by
$$
\hat{R}_{\tt reg}(g) = \frac{1}{n}\sum_{i=1}^n{\log\left[1+\exp\left\{-y_i g(\vx_i)\right\}\right]} + \frac{\lambda}{2}\|g\|_{\mathcal{H}_{\mathcal{K}}}^2
$$
where
$$
g(\vx_i) = v + \sum_{j=1}^n{w_j \mathcal{K}(\vx_i,\vx_j)}.
$$
with $g = v + h$, $v \in \Real$ and $h \in  \mathcal{H}_{\mathcal{K}}$. Here, $\mathcal{H}_{\mathcal{K}}$ is the Reproducing Kernel Hilbert Space (RKHS)
engendered by the kernel $\mathcal{K}$.  The predicted class (label) of $\vx$ via KLR is given by
$$
\hat{f}_{\tt KLR}(\vx)= {\tt sign}\left(\frac{1}{1+\exp\{-\hat{g}(\vx)\}}-\frac{1}{2}\right).
$$
One of the main advantages of KLR over SVM lies in the fact that KLR unlike SVM provides both the predicted label (hard classification) and the probability (soft classification) thereof, while SVM is inherently built to provide only the predicted label. KLR also extends naturally to multi-class while SVM requires more complicated modelling to extend beyond binary classification.

\subsection*{\it Classification and Regression Trees Learning}
Understanding trees is indeed straightforward
as they are intuitively appealing piecewise functions operating on a partitioning of the input space. Given $\mathcal{D} = \Big\{(\vx_1, Y_1),  \cdots, (\vx_n, Y_n)\Big\}$, with $\vx_i \in \mathcal{X}$, $Y_i \in \{1,\cdots,g\}$.
If $T$ denotes the tree represented by the partitioning of $\mathcal{X}$ into $q$ regions $R_1, R_2, \cdots, R_q$ such that $\mathcal{X} = \displaystyle \cup_{\ell=1}^q{R_\ell},$
then, all the observations in a given terminal node (region) will be assigned the same label, namely
$$
c_{\ell} = \underset{j \in \{1,\cdots,g\}}{\tt  argmax} \left\{\frac{1}{|R_\ell|}\sum_{\vx_i \in R_{\ell}}{I(Y_i = j)}\right\}
$$
As a result, for a new point $\vx$, its predicted class is given by
$$
\hat{Y}_{\tt Tree} = \hat{f}_{\tt Tree}(\vx) = \sum_{\ell=1}^{q}{c_{\ell}\Id_{\ell}(\vx)},
$$
where $\Id_{\ell}(\cdot)$ is the indicator function of $R_{\ell}$, i.e. $\Id_{\ell}(\vx)=1$ if $\vx \in R_{\ell}$ and $\Id_{\ell}(\vx)=0$ if $\vx \notin R_{\ell}$.
Trees are known to be notoriously unstable. Methods like bagging described earlier are often applied to
trees to help reduce the variance of the tree estimator.

\section{Predictive Performance Comparison of Learning Machines on some Massive Datasets}
When we are given a benchmark test set $\{(\vx_1, \ry_1), (\vx_2, \ry_2), \cdots, (\vx_m, \ry_m)\}$, we can assess the generalizability (predictive strength) of a regression function $f$ using the Empirical Prediction Error ({\tt EPE}) defined as
$$
{\tt EPE}(f) = \frac{1}{m}\sum_{j=1}^m{\ell(\ry_j,f(\vx_j))}.
$$
In the $k$-Neatest Neighbors context, we consider $S_k = \{k_{\tt min},\cdots,k_{\tt max}\}$, the set of possible values of $k$, and the optimal $k$ can then be estimated as
$$
\hat{k}^{(\tt opt)} = {\tt arg}\underset{k\in S_k}{\tt min}\Big\{{\tt EPE}(\hat{f}_{\tt kNN})\Big\}.
$$
It's often the case in practice that the training set is the only dataset available. In such cases, the training set is subsampled. Typically, the data set is split into training set and test set, and many realizations of the {\tt EPE} are computed over many replications of the split. Let $\hat{f}_j$ be a regression estimator and let $\hat{f}_j^{(r)}$ be its  $r$-th replication based on the $r$th split of the data into training and test sets. Now, let
$$
E_r = {\tt EPE}(\hat{f}_j^{(r)})
$$
be the $r$th replication of the test Empirical Predictive Mean Squared Error based on the test portion of the split. Then we have
$E_1, E_2, \cdots, E_R$, and can perform all kinds of statistical analyses on these numbers in order to gain deeper insights in the predictive strength of
$\hat{f}_j$. For instance, given different competing estimators $\hat{f}_1, \hat{f}_2, \cdots, \hat{f}_s$, we can plot comparative boxplots to assess virtually which of the estimators has the best performance. We could also simply compute measures of central tendency and measures of spread on these empirical predictive measures.
Clearly, this gives us a potent framework that can be used for determining the predictively optimal value of $k$ when using the kNearest Neighbors Algorithm. Indeed, if we consider each value of $k$ as defining a different regression estimator, we can then compare them using {\tt EPE} and indeed choose the value of $k$ at which
{\tt EPE} is minimized.

A more commonly used approach involving a systematic subsampling as opposed to random (stochastic) subsampling is provided by the ubiquitous cross validation tool.
To choose the optimal number of neighbors by {\it leave one out cross validation (LOOCV)}, one computes
$$
\hat{k}^{(\tt opt)} = {\tt arg}\underset{k\in S_k}{\tt min}\Big\{{\tt CV}(\hat{f}_{\tt kNN})\Big\},
$$
where $S_k = \{k_{\tt min},\cdots,k_{\tt max}\}$ is the set of possible values of $k$, and
$$
{\tt CV}(\hat{f}_{\tt kNN}) = \frac{1}{n}\sum_{i=1}^{n}{\ell(y_i,\hat{f}_{\tt kNN}^{(-i)}(\vx_i))}
=\frac{1}{n}\sum_{i=1}^{n}{1_{\{y_i\neq\hat{f}_{\tt kNN}^{(-i)}(\vx_i)\}}}
$$
with $\hat{f}_{\tt kNN}^{(-i)}(\cdot)$ denoting an instance of the estimator $\hat{f}_{\tt kNN}$ obtained without the $i$th observation.
Once the "optimal" value of $k$ is estimated using one of the above approach, we compare the predictive performance of the k-Nearest Neighbors estimates
to the estimates produced by other function estimators.

Our first data set the {\it Musk} data set, with $n=476$ and $p=166$ which will fall in category \textcolor{red}{\it E} of our taxonomy, since $n<1000$
and $1<n/p< 10$.
\begin{table}[!h]
\centering
\begin{tabular}{lrrrrrrrrr}
\toprule
   & {\sf LDA}& {\sf SVM} & {\sf CART} & {\sf rForest} & {\sf GaussPR} & {\sf kNN} & {\sf adaBoost} & {\sf NeuralNet} & {\sf Logistic}\\\hline
{\tt Musk}&    0.2227 & 0.1184 & 0.2450 & 0.1152 & 0.1511 & 0.1922 & 0.1375 & 0.1479 & 0.2408\\
{\tt Pima} &   0.2193 & 0.2362 & 0.2507 & 0.2304 & 0.2304 & 0.3094 & 0.2243 & 0.2570 & 0.2186\\
{\tt Crabs} &  0.0452 & 0.0677 & 0.1970 & 0.1097 & 0.0702 & 0.0938 & 0.1208 & 0.0350 & 0.0363\\
\bottomrule
\end{tabular}
\end{table}

\begin{figure}[!h]
\centering
\begin{tabular}{c}
\epsfig{figure=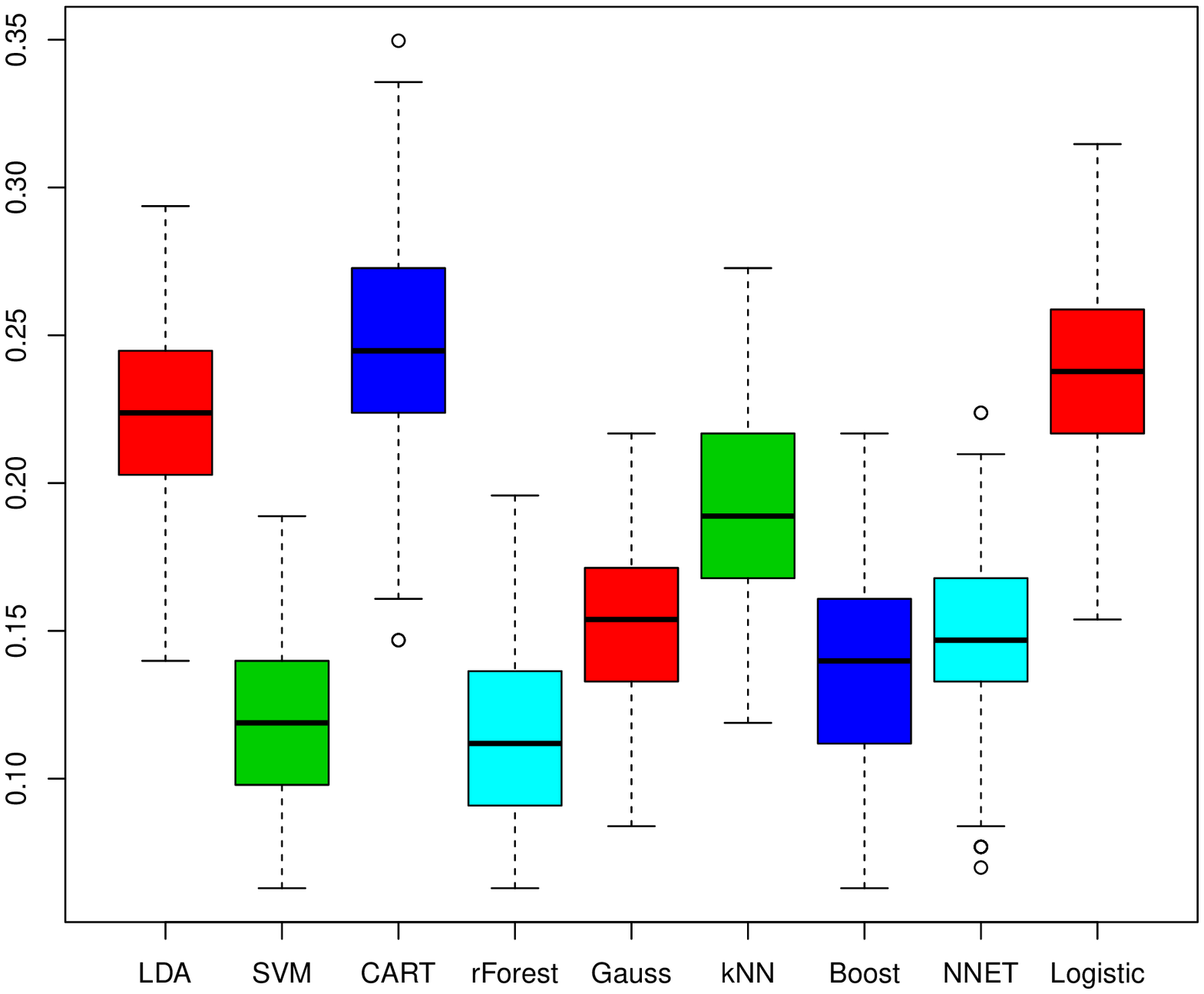, height=8cm, width=15cm}
\end{tabular}
\caption{{\it Comparison of the average prediction error over $R=100$ replications on the Musk data}}
\label{fig:jisas:musk:1}
\end{figure}

\begin{figure}[!h]
\centering
\begin{tabular}{c}
\epsfig{figure=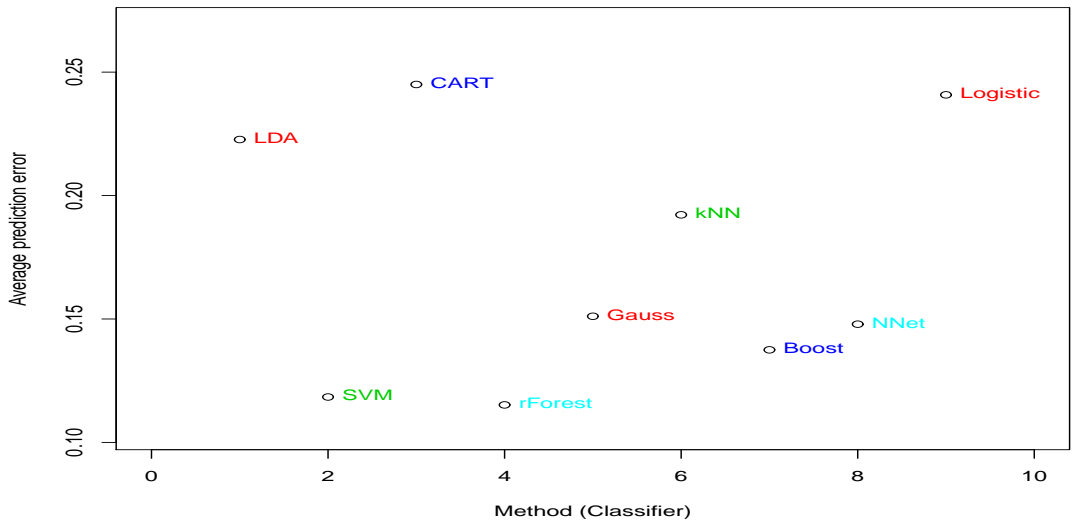, height=8cm, width=15cm}
\end{tabular}
\caption{{\it Comparison of the average prediction error over $R=100$ replications on the Musk data}}
\label{fig:jisas:musk:2}
\end{figure}

\section{Conclusion and discussion}
Throughout this paper, we have attempted to provide a detailed account of various aspects of big data. The taxonomy
provided here is neither unique nor exhaustive, but we hope to have sown the seed for a more rigorous debate
on what really does constitute big. In the interest of concision, we left out the computational exploration
of data in the case of large $p$ small $n$ because such an exploration constitutes the subject of lengthy papers
in their own right. One thing that seems to emerge in our attempt to define and develop a taxonomy of
big data is the so called No free Lunch theorem mentioned earlier. As our simulations show,
no technique appears to be universally the best on all data scenarios. Commonsense should lead us
to realize that any attempt a deriving a universally superior technique naturally
leads to unmanageably complex models that may well turn to be unusable at best and 
even inferior in performance (due to their complexity) at worst. 
We plan to explore in greater depth aspects of the taxonomy that we only scratched here, namely
diving into deeper insights on model tuning in the context of large $p$ small $n$,
the challenges of efficient and effective parallelization in the context of distributed machine learning.

\section{Acknowledgments}
I wish to express his heartfelt gratitude and infinite thanks to our Lady of Perpetual Help for Her ever-present support and guidance, especially for the uninterrupted flow of inspiration received through Her most powerful intercession. I also wish to thank Professor Boyan Dimitrov and Professor Leszek Gawarecki for inviting me to the Flint International Statistics Conference (FISC),
we also thank the reviewers for their helpful comments and suggestions that led the improvement of this article.

\bibliographystyle{chicago}
\bibliography{fokoue-papers-2013}
\end{document}